\pdfoutput=1
\documentclass[11pt]{article}

\usepackage[final]{mtsummit25}
\usepackage{copyright}

\usepackage{times}
\usepackage{latexsym}
\usepackage[T1]{fontenc}
\usepackage[utf8]{inputenc}
\usepackage{microtype}
\usepackage{inconsolata}
\usepackage{graphicx}
\usepackage{amssymb} 

\usepackage{arabtex} 
\renewenvironment{abstract}%
         {\centerline{\large\bf Abstract}%
          \begin{list}{}%
             {\setlength{\rightmargin}{0.6cm}%
              \setlength{\leftmargin}{0.6cm}}%
           \item[]\ignorespaces}%
         {\unskip\end{list}}
         
\usepackage{utf8}
\setcode{utf8}

\title{Speech-to-Speech Translation Pipelines\\for Conversations in Low-Resource Languages} 

\author{
 \textbf{Andrei Popescu-Belis\textsuperscript{1,3}},
 \textbf{Alexis Allemann\textsuperscript{1}},
 \textbf{Teo Ferrari\textsuperscript{1}},
 \textbf{Gopal Krishnamani\textsuperscript{2}}
\\
\\
 \textsuperscript{1}HEIG-VD~/~HES-SO, 1401 Yverdon-les-Bains, Switzerland\\
 \textsuperscript{2}Bhaasha Sàrl, 1400 Yverdon-les-Bains, Switzerland\\
 \textsuperscript{3}EPFL, 1015 Lausanne, Switzerland\\
 \small{
   \textbf{Correspondence:} \href{mailto:andrei.popescu-belis@heig-vd.ch}{andrei.popescu-belis@heig-vd.ch}
 }
}

\begin{document}
\maketitle
\begin{abstract}
The popularity of automatic speech-to-speech translation for human conversations is growing, but the quality varies significantly depending on the language pair.  In a context of community interpreting for low-resource languages, namely Turkish and Pashto to/from French, we collected fine-tuning and testing data, and compared systems using several automatic metrics (BLEU, COMET, and BLASER) and human assessments.  The pipelines included automatic speech recognition, machine translation, and speech synthesis, with local models and cloud-based commercial ones.  Some components have been fine-tuned on our data.  We evaluated over 60 pipelines and determined the best one for each direction.  We also found that the ranks of components are generally independent of the rest of the pipeline.
\end{abstract}

\section{Introduction} 

One of the most challenging applications of spoken language translation is real-time interpreting of human conversations.  We consider the application to community interpreting, for ethnic minorities who need assistance to access services across a language barrier, e.g., for healthcare, asylum rights, or education. 
The case study presented here involves Bhaasha, a company that provides services for community interpreting, and the Data Science group of HEIG-VD, an academic partner. Due to a growing demand, the company aims to clarify whether a system for automated interpreting meets certain quality thresholds and can be offered when human interpreters are not available.  While several online offers exist, these systems do not include the desired language pairs, or their quality is clearly insufficient, and privacy is not guaranteed.

We present the methods and the results of a joint project aimed at determining the best speech-to-speech translation pipeline made from off-the-shelf components, cloud-based services, or fine-tuned models, for two language pairs that are in high demand, but are insufficiently supported by existing systems: French-Turkish and French-Pashto.  
The paper is organized as follows.
In Section~\ref{sec:data}, we present methods for collecting and annotating data representative of the intended context of use.  In Section~\ref{sec:pipelines}, we outline the design of translation pipelines,
whose components can be smoothly interchanged.  In Section~\ref{sec:results}, we evaluate all combinations of four ASR, three MT, and two TTS components, either local or cloud-based, also including two fine-tuned components and a speech-to-text translation one.  We present evaluation scores from automatic metrics and determine the best combination of components per direction.  We also show that the ranking of components is generally independent of the other modules in a pipeline.  Finally, we present human scores over a subset of the data, showing that accuracy, fluency and intonation of the best pipelines are considered as `good' or `very good'.


\section{State of the Art}
\label{sec:sota}

Methods for speech-to-\textit{text} translation (e.g.\ for subtitling) have been the subject of many recent publications, unlike methods for speech-to-\textit{speech} translation, as the speech synthesis part is difficult to train.  Moreover, spoken translation has been studied more often for monologues than for conversations.  The three necessary components are automatic speech recognition (ASR, or speech-to-text, STT), machine translation (MT), and speech synthesis (or text-to-speech, TTS).

Research interests, however, have shifted from loosely coupled cascades of ASR and MT, to tighter coupling, and finally to recent end-to-end models \citep{sperber-paulik-2020-speech,xu2023recentadvancesdirectspeechtotext}. 
For instance, an approach to multilingual speech-to-text translation through efficient transfer learning from a pretrained speech encoder (wav2vec) and text decoder (BERT), is proposed by \citet{li-etal-2021-multilingual}.  \citet{dong2021listen} propose a Listen, Understand and Translate (LUT) approach to train end-to-end speech-to-text translation.

\citet{bentivogli-etal-2021-cascade} compare the two paradigms -- cascaded vs.\ end-to-end -- and claim that the gap between them is almost closed.  However, for low-resource languages, end-to-end systems are difficult to train due to the lack of data, while cascaded systems can use components trained with simpler tasks.  Alternatively, massively multilingual systems such as Whisper \citep{radford2022whisper} for ASR~+ MT claim that low-resource languages are improved thanks to higher-resource similar languages.  For instance, cascaded approaches can take advantage of optimized low-resource MT components \citep{alex-r-atrio-popescu-belis-2022-interaction}.

The IWSLT 2022, 2023 and 2024 evaluation campaigns \citep{anastasopoulos-etal-2022-findings-shortened,agrawal-etal-2023-findings-shortened,ahmad-etal-2024-findings-shortened} featured various shared tasks, including speech-to-text and speech-to-speech translation for low-resource languages. A typical low-resource system presented at IWSLT 2023 is the Marathi to Hindi submission by \citet{kesiraju-etal-2023-systems}, including an end-to-end and a cascaded system.  Various techniques for improving low-resource speech-to-text translation, in particular with initialization from a multilingual ASR system, have been proposed \citep{khosravani2021learning,fu2023improving,Kesiraju_2023}.  

Large corpora exist for well-resourced language pairs, but not for the low-resource ones that we target.  While datasets of recorded speech can be more easily found, datasets with transcriptions and translations are scarce or inexistent.  The MuST-C \citep{di-gangi-etal-2019-must} and Multilingual TEDx corpora include speech and translation in English and 8 other European languages, but not Turkish or Pastho.  CoVoST-2 \citep{wang2021covost} covers speech translation from several languages to/from English and includes Turkish.  This resource was used to test the Whisper ASR~+ MT used here.

Recent developments of generative AI and large language models have enabled significant progress in speech translation and synthesis, but low-resource languages are still insufficiently supported.  For instance, several companies advertise multilingual speech translation systems on the Web, as apps for smartphones, or as cloud-based services, mostly for for a few well-resourced languages.  For instance, one of the major players, Google, offers the three components individually via APIs, but also bundles them into a pipeline that often appears in informal tests as one of the best translation apps for several language pairs.  Other commercial offers include DeepL, Microsoft's Bing Translator, iTranslate, SayHi, Translate Now, Yandex, or Talking Translator.  Many of the related apps have received reviews from their users, which provide a form of evaluation, although ratings for specific language pairs are rarely found.  In our tests, we observed that these solutions are not ready for the low-resource languages studied here, nor for use in the setting of community interpreting.


\begin{table*}[ht] 
  \centering 
    \begin{tabular}{|l|l|l|l|l|}
    \hline
        dial. & utt. & lang & audio & transcription \\ \hline
        B001 & 1 & fr & B001-1-fr.wav & Bonjour Monsieur, qu’est-ce qui vous amène ? \\ \hline
        B001 & 1 & ps & B001-1-ps.wav & \RL{سلام ښاغلی، تاسو دلته څنګه راغلي ياست؟} \\ \hline
        B001 & 1 & tr & B001-1-tr.wav & Merhabalar beyefendi, bugün neden buradasınız? \\ \hline
        B001 & 2 & fr & B001-2-fr.wav & J’ai mal à la tête, très mal depuis déjà plus de deux semaines. \\ \hline
        B001 & 2 & ps & B001-2-ps.wav & \RL{زه له دوو اونیو راهیسې ډیر بد سر درد لرم} \\ \hline
        B001 & 2 & tr & B001-2-tr.wav & 2 haftadan fazla başım ağrıyor. \\ \hline
        B001 & 3 & fr & B001-3-fr.wav & Qu’est-ce que vous prenez pour calmer ces douleurs ? \\ \hline
        B001 & 3 & ps & B001-3-ps.wav & \RL{تاسو د دې درد څخه د خلاصون لپاره څه اخلي؟} \\ \hline
        B001 & 3 & tr & B001-3-tr.wav & Ağrılar geçsin diye ne alıyorsunuz? \\ \hline
    \end{tabular}
  \caption{Three utterances from our dataset: each one is available in three languages and two modalities.}
  \label{tab:extrait-db}
\end{table*}

\section{Data Gathering and Formatting}
\label{sec:data}

To the best of our knowledge, there are no datasets with parallel conversational speech (i.e.\ interpreted in both directions) for Turkish-French and Pashto-French (tr-fr, ps-fr).  Therefore, we collected new data which suits our project's needs.

The central idea of our parallel dataset is to include complete dialogues in situations encountered by Bhaasha's community interpreters.  For each utterance, we have a reference transcript and an audio recording, in each of the three languages of the project: an excerpt is shown in Table~\ref{tab:extrait-db}.  For each utterance, in each language, the dataset contains indexing information (dialogue codename, utterance number, and language), the transcript of the utterance, the name of the audio file with the utterance (similarly indexed), 
and whether it is used in the fine-tuning or the testing subsets. 
With this structure, the dataset can be used to fine-tune or to test speech translation pipelines in any translation direction.

\subsection{Data Sources}

Collecting such a dataset requires an abstraction over the complex reality of community interpreting, which involves three speakers: the two persons between whom the dialogue takes place, and the interpreter, who interprets consecutively the speech in both directions.  However, it appeared early in the project that real dialogues mediated by interpreters \emph{could not be recorded due to privacy reasons.} 

Therefore, for most of our data, we settled on the following protocol.  We gathered or wrote dialogues similar to those handled by Bhaasha interpreters, writing them in French, in some cases with the help of the GPT-4 LLM.  Then, we asked interpreters from Bhaasha to write translations of the entire dialogues into Turkish or Pachto, by postediting automatic translations from the Google or Microsoft online systems.  Finally, we recorded interpreters reading aloud these translations, and added French audios read by different native speakers, manually segmenting all audios into utterances.  

This text-centric protocol appeared to be much more efficient than an audio-based one in which interpreters listen to a source sentence (or read a sentence) and then utter the spoken translation in the target language, which is recorded and then transcribed.
This solution was very demanding for interpreters, 
and was not entirely natural as it required interpreting both dialogue participants in the same direction. 
Enacting original new spoken dialogues appeared also to have too high transcription and translation costs.
The current dialogues, although more fluent than real ones, are the best substitute that could be found within the frame of our project.

The sources of the dialogues included in our project data are the following ones.  Each dialogue is identified by a letter coding the generation method and an index number.  
Each separate sentence (utterance) appears on one line, and speaker turns can be made of one or more lines (as indicated in the metadata, see~\ref{sec:exchange-format}). 

\begin{itemize} \setlength{\itemsep}{-1pt}
    \item `G' series (G001--G013, 630 lines): dialogues generated with GPT-4.\footnote{\url{https://chat.openai.com}}  This was the quickest technique and provided about half of our data.  Given a precise prompt in French,\footnote{Prompts describe in detail a situation, matching closely those encountered by interpreters, e.g., ``Write a dialogue at the welfare office with this topic: a young man has found a part-time job (50\%) and wants to know what impact this will have on his welfare.  He will have a long and costly commute. Generate 30 to 40 turns.''} GPT-4 generated a realistic in-domain dialogue of the desired size and style, which was improved to satisfactory levels with minimal human edits from the experimenters. 
    \item `C' series (C001--C004, 498 lines): excerpts from the CoVoST-2 corpus \citep{wang2021covost}, Turkish-English parallel subset, mostly with spoken news (admittedly, not dialogues).  We translated the data into French by post-editing MT output, and added French audio from native speakers.
    \item `B' series (B001--B006, 180 lines): dialogues created as French text by interpreters from Bhaasha, in the spirit of those that they encounter as community interpreters.
    \item `P' series (P001--P004, 110 lines): four samples of learners' material in French, corresponding to our style and topics.
\end{itemize}

\begin{table}[h]
  \centering
  \begin{tabular}{c|rr|rrr}
    \hline
    & \multicolumn{2}{c|}{\bf Utterances} & \multicolumn{3}{c}{\bf Durations} \\
    & \textbf{train} & \textbf{test} & \textbf{sec.} & \textbf{min.} & \textbf{words} \\\hline
    fr    & 722 & 723 & 4,138 & 69 & 11,986 \\
    tr    & 722 & 723 & 4,985 & 83 & 8,295 \\
    ps    & 0 & 400   & 1,741 & 29 & 3,717 \\\hline
    total  & \multicolumn{2}{|c|}{3,290} & 10,864 & 181 & 23,998  \\\hline
  \end{tabular}
  \caption{Data used for fine-tuning and testing.}
  \label{tab:datasizes}
\end{table}

As summarized in Table~\ref{tab:datasizes}, our dataset includes 28 dialogues with 1,445 utterances (lines).  All utterances are available in French and Turkish, with transcript and audio, but Pashto translation is partial, from lack of availability of Pashto interpreters.  We randomly sampled 723 lines for testing and 722 for fine-tuning from the French side, and similarly from the Turkish side.  We did not sample entire dialogues, to ensure better similarity between fine-tuning and testing data.  For Pashto, all 400 lines were used for testing.

\subsection{Exchange Format}
\label{sec:exchange-format}

The exchange format is kept simple, to ensure easy reuse.  The dataset is contained in two text-based files and one folder with audio files:
\begin{itemize} \setlength{\itemsep}{-1pt}
    \item \texttt{dialogues.json} -- metadata in JSON format, described below.
    \item \texttt{dataset.csv} -- indexed transcripts and names of audio files, as shown in Table~\ref{tab:extrait-db}.
    \item \texttt{audios} -- contains one audio file per utterance, named using indexes, from recordings on smartphones or laptops in silent environments (2 channels, 48 kHz, 32 bits).
\end{itemize}

As metadata, we include for each dialogue identified by its codename: long name or brief description, creation method (including prompt to GPT-4 for the G series), date of recording, and number of utterances.  For each language, we indicate whether it is an original or translated version, how it was translated, and the total duration of audios.  Finally, we indicate the grouping of utterances in speaker turns using their index numbers.


\begin{table*}
  \centering
  \begin{tabular}{lllll}
    \hline
\textbf{Stage} & \textbf{Type} & \textbf{Name} & \textbf{URL} \\
\hline
    & cloud & Google STT &  \small\url{https://cloud.google.com/speech-to-text/v2} \\
    & cloud & Microsoft STT & \small\url{https://speech.microsoft.com/portal} \\ 
ASR & local & OpenAI Whisper-large-v3 & \small\url{https://huggingface.co/openai/whisper-large-v3} \\
    & local & Fine-tuned Whisper & \small\url{https://huggingface.co/openai/whisper-large-v3} \\ 
    & local & Meta MMS & \small\url{https://huggingface.co/facebook/mms-1b-all} \\
\hline
\it ASR+MT & \it local & \it Whisper Translate & \small\url{https://huggingface.co/openai/whisper-large-v3} \\
\hline
    & cloud & Google MT & \small\url{https://cloud.google.com/translate} \\
    & cloud & Microsoft MT & \small\url{https://www.microsoft.com/en-us/translator}\\
MT  & local & NLLB-200 3.3B & \small\url{https://huggingface.co/facebook/nllb-200-3.3B}\\
    & local & Fine-tuned NLLB (1.3B) & \small\url{https://huggingface.co/facebook/nllb-200-1.3B}\\
    & \it local & \it HelsinkiNLP & \small\url{https://huggingface.co/Helsinki-NLP}\\ 
\hline
    & cloud & Google TTS & \small\url{https://cloud.google.com/text-to-speech}\\
TTS & cloud & Microsoft TTS & \small\url{ https://speech.microsoft.com/portal}\\
    & \it local & \it YourTTS & \small\url{https://github.com/Edresson/YourTTS}\\
\hline
  \end{tabular}
  \caption{Components used in our speech-to-speech translation pipelines (in italics, preliminary studies only).}
  \label{tab:components}
\end{table*}

\section{Speech-to-Speech Translation Pipelines}
\label{sec:pipelines}

\subsection{Components}

We considered all possible combinations of the following ASR, MT and TTS components.  Table~\ref{tab:components} below provides the exact names and URLs of all of them.
We evaluated ASR/MT/TTS cloud-based commercial components from Google and Microsoft, as well as the following open-weight models run locally.  For ASR, we tested Whisper from OpenAI \citep{radford2022whisper} in `transcribe' mode, i.e.\ in the same language, and MMS from Meta \citep{pratap2023mms}.  For MT, we tested the multilingual NLLB-200 model with 3.3B parameters \citep{nllbteam2022languageleftbehindscaling}.  But for TTS, no competitive local model could be found for our languages.  Moreover, we fine-tuned Whisper and NLLB-200 with 1.3B parameters on the training subset of the fr-tr data (see Table~\ref{tab:datasizes}), resulting in the models prefixed with `ft' below.

\subsection{Architecture}

We built a flexible application to support experimentation, but also real-time demonstration.  Hence, the application includes a frontend and a backend, and is hosted on the Kubernetes infrastructure of the Swiss AI Center\footnote{\url{https://swiss-ai-center.ch}} with S3 MinIO storage.  The dataset is managed using DVC.

The frontend of the application is developed using the React framework, while the backend is built in Python with FastAPI, providing several HTTP endpoints to enable the use of different versions of the ASR, MT, and ST modules.  The frontend orchestrates the sequence of calls across the various stages of the speech-to-speech translation pipeline.  These endpoints allow responses to be generated using either local models running on GPUs or remote models accessed via third-party APIs.  Additionally, for every request, the backend stores copies of the audio and model outputs at each stage in a S3 bucket, which facilitates analysis and human evaluation.

A frontend interface allows human users to inspect or demonstrate the system.
The interface enables on-the-fly change of components in the pipeline, depending on the desired source and target languages.  A laptop with a regular microphone can be used for demos.

\subsection{Evaluation Metrics}

We use Word Error Rate (WER) to score ASR components, with the JiWER Python package.\footnote{\url{https://github.com/jitsi/jiwer}}
We use four automatic metrics for MT: three of them, available from the Sacrebleu library \citep{post-2018-call},\footnote{\url{https://github.com/mjpost/sacrebleu}} use various form of edit distance between candidate and reference translations: BLEU, ChrF, and Translation Error Rate (TER). 
The fourth metric, COMET \citep{rei-etal-2022-comet}\footnote{\url{https://github.com/Unbabel/COMET}}, compares source and target embeddings using a large language model (wmt22-comet-da), and is applicable to French-Turkish as well as French-Pashto.  We found that there is a strong correlations between these metrics: using each system as a data point, average pairwise Pearson correlation is 0.89 for fr-tr and 0.97 for fr-ps, with four metrics.  Therefore, we use below two representative and least correlated metrics, namely BLEU and COMET.


\section{Results}
\label{sec:results}

\subsection{ASR Scores (WER)}

The WER scores for the ASR components are given in Table~\ref{tab:fr-tr-ps-ASR-WER} (lower is better).  The rankings are consistent across French and Turkish, although the differences between systems are not.  For Turkish, the fine-tuning of Whisper on our data brings a visible improvement (from 0.14 to 0.09), while the untuned Whisper performs on par with the Microsoft cloud-based service.  The Google service and the Meta local model follow at some distance.

\begin{table}[ht]
  \centering
  \begin{tabular}{l|c}
    \hline
    \textbf{French ASR}      & \textbf{WER} \\\hline
    ft\_whisper\_transcribe  & \textbf{0.04} \\
    whisper\_transcribe      & 0.06 \\
    microsoft\_stt           & 0.08 \\
    google\_stt              & 0.23 \\
    meta\_mms                & 0.24 \\\hline
    \textbf{Turkish ASR}      & \\\hline
    ft\_whisper\_transcribe  & \textbf{0.09} \\
    whisper\_transcribe      & 0.14 \\
    microsoft\_stt           & 0.15 \\
    google\_stt              & 0.31 \\
    meta\_mms                & 0.40 \\\hline
    \textbf{Pashto ASR}      & \\\hline   
    microsoft\_stt           & \textbf{0.45} \\
    google\_stt              & 0.89 \\
    whisper\_transcribe      & 0.92 \\\hline
  \end{tabular}
  \caption{WER for French, Turkish, and Pashto.}
  \label{tab:fr-tr-ps-ASR-WER}
\end{table}

\subsection{MT Scores (BLEU and COMET)}

The scores of written MT for Turkish and French (both directions) for all combinations of modules, with four metrics, are shown in Table~\ref{tab:fr-and-tr-raw-MT} in the Appendix.  Similarly, the MT scores for Pashto and French are shown in Table~\ref{tab:fr-and-ps-raw-MT}.   These tables indicate the best ASR+MT pipelines, with substantial agreement between metrics:
\begin{itemize} \setlength{\itemsep}{-1pt}
    \item \textbf{tr-fr:} fine-tuned Whisper (or not fine-tuned) with Google MT (or Microsoft MT).
    \item \textbf{fr-tr:} fine-tuned Whisper (or not fine-tuned) with Microsoft MT (or Google MT).
    \item \textbf{ps-fr:} Microsoft ASR with Microsft MT (or Microsoft MT).
    \item \textbf{fr-ps:} fine-tuned Whisper (or not fine-tuned) with Google MT.
\end{itemize}
To perform a systematic analysis of the intrinsic quality of each module and of the effects of their combinations, we propose the following approach, applied to each translation direction.  

\subsubsection{Turkish and French}

\begin{table}[ht]
\setlength{\tabcolsep}{3pt}
  \centering
    \begin{tabular}{ll|c|c}
    \hline
        \textbf{ASR} & \textbf{MT} & \textbf{COMET} & \textbf{AVG} \\ \hline 
        whisper  & google\_mt  & \textbf{89.60} & 87.47 \\ 
        ft\_whisper  & google\_mt  & 89.40 & ~ \\ 
        microsoft\_stt  & google\_mt  & 87.87 & ~ \\
        google\_stt  & google\_mt  & 87.28 & ~ \\ 
        meta\_mms  & google\_mt  & 83.20 & ~ \\ 
        \hline
        whisper  & microsoft\_mt  & 88.42 & 86.04 \\ 
        ft\_whisper  & microsoft\_mt  & 88.06 & ~ \\ 
        microsoft\_stt  & microsoft\_mt  & 86.64 & ~ \\ 
        google\_stt  & microsoft\_mt  & 86.49 & ~ \\ 
        meta\_mms  & microsoft\_mt  & 80.61 & ~ \\ 
        \hline
        whisper  & ft\_nllb-1.3B  & 86.14 & 83.86 \\ 
        ft\_whisper  & ft\_nllb-1.3B  & 85.30 & ~ \\ 
        microsoft\_stt  & ft\_nllb-1.3B  & 84.21 & ~ \\ 
        google\_stt  & ft\_nllb-1.3B  & 84.21 & ~ \\ 
        meta\_mms  & ft\_nllb-1.3B  & 79.42 & ~ \\ 
        \hline
        whisper  & nllb-3.3B  & 85.92 & 83.62 \\ 
        ft\_whisper  & nllb-3.3B  & 84.99 & ~ \\ 
        microsoft\_stt  & nllb-3.3B  & 84.02 & ~ \\ 
        google\_stt  & nllb-3.3B  & 83.81 & ~ \\
        meta\_mms  & nllb-3.3B  & 79.38 & ~ \\ \hline
    \end{tabular}
  \caption{COMET scores for Turkish-to-French speech-to-text translation, grouped by MT system, and ranked by average COMET over each group.}
  \label{tab:tr-fr-byMT-COMET}
\end{table}

We first present a detailed analysis of \textbf{Turkish $\rightarrow$ French} pipelines, and then summarize conclusions for the other direction, and then for Pashto and French.  
We organize the COMET scores in two ways.  
First, as shown in Table~\ref{tab:tr-fr-byMT-COMET}, pipelines are grouped by MT systems, and the groups are ranked by average COMET.  Inside each group, ASR components are ranked too.  We find that the ranking of ASR is the same inside the first and second best groups, which are those with Google MT and Microsoft MT (with a large difference between them).  The ranking of the first two ASR systems is permuted when we move to the third and fourth groups.  Therefore, the following stable ranking is found for Turkish ASR.  Fine-tuning Whisper turns out to be beneficial to BLEU scores but not to COMET ones.
\begin{quote}
\texttt{Whisper $>$ Fine-tuned Whisper $>$ Microsoft ASR $>$ Google ASR $>$ Meta MMS ASR}
\end{quote}

\begin{table}[ht]
\setlength{\tabcolsep}{3pt}
  \centering
    \begin{tabular}{ll|c|c}
     \hline
        \textbf{ASR} & \textbf{MT} & \textbf{COMET} &  \textbf{AVG} \\ 
        \hline
        whisper  & google\_mt  & \textbf{89.60} & 87.52 \\ 
        whisper  & microsoft\_mt  & 88.42 & ~ \\ 
        whisper  & ft\_nllb-1.3B  & 86.14 & ~ \\
        whisper  & nllb-3.3B  & 85.92 & ~ \\ 
        \hline
        ft\_whisper  & google\_mt  & 89.40 & 86.94 \\ 
        ft\_whisper  & microsoft\_mt  & 88.06 & ~ \\ 
        ft\_whisper  & ft\_nllb-1.3B  & 85.30 & ~ \\ \
        ft\_whisper  & nllb-3.3B  & 84.99 & ~ \\ 
        \hline
        microsoft\_stt  & google\_mt  & 87.87 & 85.69 \\ 
        microsoft\_stt  & microsoft\_mt  & 86.64 & ~ \\ 
        microsoft\_stt  & ft\_nllb-1.3B  & 84.21 & ~ \\ 
        microsoft\_stt  & nllb-3.3B  & 84.02 & ~ \\ 
        \hline
        google\_stt  & google\_mt  & 87.28 & 85.45 \\ 
        google\_stt  & ft\_nllb-1.3B  & 86.49 & ~ \\ 
        google\_stt  & microsoft\_mt  & 84.21 & ~ \\ 
        google\_stt  & nllb-3.3B  & 83.81 & ~ \\ 
        \hline
        meta\_mms  & google\_mt  & 83.20 & 80.65 \\ 
        meta\_mms  & microsoft\_mt  & 80.61 & ~ \\ 
        meta\_mms  & ft\_nllb-1.3B  & 79.42 & ~ \\ 
        meta\_mms  & nllb-3.3B  & 79.38 & ~ \\ 
        \hline
    \end{tabular}
  \caption{COMET scores for Turkish-to-French speech-to-text translation, grouped by ASR system, and ranked by average BLEU over each group.}
  \label{tab:tr-fr-byASR-COMET}
\end{table}

Second, as shown in Table~\ref{tab:tr-fr-byASR-COMET}, pipelines are grouped by ASR system, and the groups are ranked by average COMET; inside each group, MT components are ranked too.  We find that the ranking of MT is almost always the following one (except in one group where the second and third ranks are permuted):
\begin{quote}
\texttt{Google MT $>$ Microsoft MT $>$ Fine- tuned NLLB 1.3B $>$ NLLB 3.3B}
\end{quote}

For the \textbf{French $\rightarrow$ Turkish pipelines}, a similar analysis shows that the stable ranking of ASR components in each grouping based on MT is the following one (the ranking of the last two components is reversed in half of the groups):
\begin{quote}
\texttt{Fine-tuned Whisper $>$ Whisper $>$ Microsoft ASR $>$ Google ASR $\approx$ Meta MMS ASR}
\end{quote}
Conversely, when grouping by ASR, fine-tuned Whisper ahead of the others in BLEU score, but the untuned Whisper is slightly ahead on COMET.  They are followed by Microsoft ASR, and then at some distance by Meta MMS and Google ASR, which are quit close.  When grouping by ASR, the stable ranking of MT components is the following one, with some uncertainty over the fine-tuned NLLB, and a reversal of the first two ranks with COMET:
\begin{quote}
\texttt{Microsoft MT $>$ Google MT $>$ Fine-tuned NLLB 1.3B $>$ NLLB 3.3B}
\end{quote}

\subsubsection{Pashto and French}

Similar to the above strategy, the scores for the \textbf{Pashto $\rightarrow$ French} pipelines are either grouped by MT systems to observe the rankings of ASR in each group, or grouped by ASR systems to observe the rankings of MT.  The actual scores are shown in Table~\ref{tab:fr-and-ps-raw-MT} in the Appendix.
We make the following observations using COMET scores.  
When grouping by MT system (Google or Microsoft), the ranking of ASR is always the same:
\texttt{Microsoft ASR $>$ Whisper $>$ Google ASR}.
In terms of the actual average score per ASR, Microsoft ASR is much better than Whisper or Google ASR, which do not seem usable here.
When grouping by ASR, the ranking of MT is the same for the first two ASR systems, but is reversed for the last one, likely due to the poor quality of input to MT: \texttt{Google MT $>$ Microsoft MT}.   In terms of average per MT, Google MT is also slightly ahead of Microsoft MT, as in the observations grouped per ASR.

Finally, for the \textbf{French $\rightarrow$ Pashto} pipelines, when grouping by MT,  the rankings of ASR differ, although the best system is the Fine-tuned Whisper in both cases.  For Google MT, Whisper is the second best, although it is ranked fourth when using Microsoft MT.  However, given the poor quality of this last MT system, the rankings may not be reliable.  In terms of average per ASR, the Fine-tuned Whisper is first, followed by Whisper and by Microsoft ASR, and then by Meta MMS and Google ASR.  (As they concern French ASR, these rankings are similar to those for fr-tr.)
When grouping per ASR, the ranking of MT is always the same: \texttt{Google MT $>$ Microsoft MT}, with large differences between the two (8--9 COMET points).


\subsection{End-to-end Scores (BLASER)}

The BLASER 2.0 scores \citep{dale-costa-jussa-2024-blaser} of the speech-to-speech translation pipelines are given in Table~\ref{tab:blaser-and-human-scores}.  They were computed for Turkish and French, as no models are available for Pashto.  We selected two representative ASR~+ MT pipelines: Whisper~+ NLLB is entirely local and not fine-tuned, while Google~+ Google is the commercial cloud-based offer from Google.  We combined each of them with two cloud-based speech synthesis solutions, respectively from Google and Microsoft, as no local TTS was satisfactory,  We computed BlaserQE and BlaserRef scores for each of the four pipelines.  For each sentence, BlaserQE compares the embeddings of the \textit{source} and of the \textit{candidate} translation in the audio modality, while BlaserRef also considers the embedding of the \textit{written reference} translation. 

{\begin{table}[ht] 
  \centering 
  \scalebox{1}{
      \begin{tabular}{|l|c|c|c|c|}
         \hline
            ASR+MT  & \multicolumn{2}{c|}{\small Whisper+NLLB}  & \multicolumn{2}{c|}{\small  Google+Google} \\ \hline
            TTS & \small  Google & \small  MS & \small  Google & \small MS \\ \hline
            \hline
            \textbf{} & \multicolumn{4}{c|}{\bf fr-tr systems} \\ \hline
            WER  & \multicolumn{2}{c|}{0.06}  & \multicolumn{2}{c|}{0.23} \\ \hline
            BLEU  & \multicolumn{2}{c|}{25.76}  & \multicolumn{2}{c|}{22.72} \\ \hline
            COMET  & \multicolumn{2}{c|}{89.37}  & \multicolumn{2}{c|}{87.49} \\ \hline
            BlaserQE  & 3.07 & 3.02 & 3.06 & 3.01 \\ \hline
            BlaserRef  & 3.24 & 3.16 & 3.25 & 3.18 \\ \hline
            Meaning  & \multicolumn{2}{c|}{4.43}  & \multicolumn{2}{c|}{4.27} \\ \hline
            Correctness  & \multicolumn{2}{c|}{4.55}  & \multicolumn{2}{c|}{4.69} \\ \hline
            Intonation  & 4.55 & 4.55 & 4.50 & 4.57 \\ \hline
             & \multicolumn{4}{c|}{\bf tr-fr systems} \\ \hline
            WER & \multicolumn{2}{c|}{0.14}  & \multicolumn{2}{c|}{0.31} \\ \hline
            BLEU  & \multicolumn{2}{c|}{38.46}  & \multicolumn{2}{c|}{43.43} \\ \hline
            COMET  & \multicolumn{2}{c|}{85.92}  & \multicolumn{2}{c|}{87.26} \\ \hline
            BlaserQE  & 3.18 & 3.16 & 3.28 & 3.26 \\ \hline
            BlaserRef  & 3.19 & 3.18 & 3.34 & 3.32 \\ \hline
        \end{tabular}
    }
  \caption{Results of automatic evaluation with BLASER 2.0 and of human evaluation of speech-to-speech translation.  For comparison purposes, we reproduce the WER, BLEU and COMET scores. MS stands for the Microsoft speech synthesis component.}
  \label{tab:blaser-and-human-scores}
\end{table}}

The BLASER 2.0 scores indicate that using Google TTS is \textit{always} slightly better than Microsoft TTS.  The difference between these systems for fr-tr is statistically significant at the 1\% level (as measured by a t-test) with the BlaserRef metric, regardless of the ASR~+ MT part.  As for tr-fr, the difference is significant at the 1\% level (t-test) only when combined with Google ASR~+ MT.  Moreover, the Google-only pipeline scores significantly better than both local ones.

\subsection{Human Evaluation}
\label{sec:validation}

As a pilot experiment, we showed 21 utterances to two human judges, native speakers of Turkish, one of them being an interpreter.
We presented them with source audio and translations from French to Turkish by the same four pipelines as in the previous section.  For each utterance, they were asked to grade three aspects: (1)~how well the original meaning is communicated by the translation; (2)~how correct is the wording of the translation; and (3)~how good is the intonation of the translation.  The first two aspects are akin to the traditional \textit{adequacy} and \textit{fluency} dimensions, but here no transcript is seen by evaluators.  The third one is aimed specifically at speech synthesis.  To speed up evaluation, when the ASR~+ MT pipeline is the same but the TTS is different, we ask evaluators to rate only once the meaning and correctness, and to rate separately the two different TTS outputs.  At the top of the interface, which includes links to the audios and a drop-down menu for each rating, we briefly defined each aspect.  The possible values for ratings are the following ones (originally in French):
\begin{itemize} \setlength{\itemsep}{-1pt}
    \item Meaning: (1)~not at all; (2)~the general idea; (3)~some elements; (4)~almost entirely; (5)~entirely.
    \item Correctness: (1)~very incorrect; (2)~quite incorrect; (3)~medium; (4)~quite correct; (5)~very correct.
    \item Intonation: (1)~not understandable; (2)~a little understandable; (3)~medium; (4)~well understandable; (5)~perfectly understandable.
\end{itemize}

Average ratings for each aspect by the two judges are given in Table~\ref{tab:blaser-and-human-scores}.  The estimated quality  by the human judge is overall between 4 and 5 for all aspects and systems.  Communicated meaning is scored around 4, i.e.\ `almost entirely', which is the lowest of the three scores, likely due to the combination of errors from ASR and MT.  Grammatical correctness, depending almost exclusively on the ASR~+ MT pipeline, is also between `quite correct' and `very correct', here with a slight advantage to the Google components (4.9 vs.\ 4.5).  This could be due to NLLB being a multilingual MT system, which has a lower fluency for Turkish than the Google's dedicated system.  Intonation, either generated by Google TTS or by Microsoft TTS, scores close to 4.5, i.e.\ between `well' and `perfectly understandable'.  There is no significant difference between the two systems, despite a slightly higher BLASER score for Google TTS.  The human ratings give an idea of the calibration of automatic metrics, with BLEU scores of around 25 and COMET scores of nearly 90 being already perceived as good quality.


\section{Conclusion}
\label{sec:conclusion}

We have produced data and assembled numerous speech-to-speech translation pipelines, for interpreting Turkish $\leftrightarrow$ French and Pashto $\leftrightarrow$ French conversations.  Specifically, we have produced three hours of data in settings compatible with community interpreting, and used half of it for fine-tuning two Turkish $\leftrightarrow$ French ASR and MT systems, and the other half for evaluation.  We scored over 60 pipelines of ASR, MT and TTS systems, either based on open-weight models run locally, or on commercial cloud-based services.  We identified the best-performing pipeline in each direction, and found that the ranking of components was consistent, regardless of the other components of pipelines.  We used four automatic evaluation metrics (WER, BLEU, COMET and BLASER), along with pilot human evaluations.  The implementation of an online system with a push-to-talk interface, along with an offline version allowing batch processing, now paves the way towards usability testing of automatic interpretation solutions, which will also need to  take into consideration factors such as privacy, cost, and deployment strategy.

\section*{Acknowledgments} 
We thank the HES-SO for support through the INTERCOM project (AGP n.\ 130496) and the Swiss AI Center (\url{https://swiss-ai-center.ch}) for support with the demonstrator in their Core Engine.
We are grateful to Dr.\ Selin Ata\c{c} from HEIG-VD, Egeas Papadopoulos from EPFL and two interpreters from Bhaasha for their help with data collection and annotation, and to Prof.\ Bertil Chapuis from HEIG-VD for the initial project at the Swiss AI Center.
We thank the three anonymous MT Summit reviewers for their valuable feedback.

\bibliography{anthology,custom}

\appendix

\begin{table*}[ht] 
\setlength{\tabcolsep}{4pt}
    \centering
    \begin{tabular}{ll|cccc|cccc}
    \multicolumn{10}{l}{\textbf{\large A~~~~~Appendix}} \\ 
    \multicolumn{10}{l}{~~} \\
    \hline
         \textbf{} & \textbf{} & \multicolumn{4}{c|}{\bf tr-fr} & \multicolumn{4}{c}{\bf fr-tr} \\  \hline
         \textbf{ASR} &  \textbf{MT} &  \textbf{BLEU} &  \textbf{ChrF} &  \textbf{TER} &  \textbf{COMET} &  \textbf{BLEU} &  \textbf{ChrF} &  \textbf{TER} &  \textbf{COMET} \\ \hline
        ft\_whisper  & ft\_nllb-1.3B  & 39.79 & 62.13 & 52.97 & 85.30 & 30.77 & 60.22 & 54.49 & 89.80 \\ \hline
        ft\_whisper  & google\_mt  & \textbf{55.78} & \textbf{72.88} & \textbf{37.05} & \textbf{89.40} & \textbf{36.93} & \textbf{65.19} & \textbf{50.31} & \textbf{90.95} \\ \hline
        ft\_whisper  & microsoft\_mt  & \textit{45.38} & \textit{67.66} & \textit{43.53} & \textit{88.06} & \textbf{38.11} & \textbf{65.51} & \textbf{49.40} & \textit{90.20} \\ \hline
        ft\_whisper  & nllb-3.3B  & 38.06 & 58.95 & 54.54 & 84.99 & 26.99 & 57.63 & 59.98 & 89.24 \\ \hline
        google\_stt  & ft\_nllb-1.3B  & 35.77 & 58.63 & 54.62 & 85.30 & 22.38 & 54.37 & 62.61 & 87.10 \\ \hline
        google\_stt  & google\_mt  & 43.43 & 66.49 & 46.32 & 87.28 & 22.72 & 58.69 & 62.12 & 87.49 \\ \hline
        google\_stt  & microsoft\_mt  & 37.14 & 63.37 & 51.04 & 86.49 & 26.24 & 60.55 & 60.98 & 88.26 \\ \hline
        google\_stt  & nllb-3.3B  & 34.65 & 55.96 & 57.39 & 83.81 & 20.06 & 53.33 & 65.61 & 87.24 \\ \hline
        microsoft\_stt  & ft\_nllb-1.3B  & 38.46 & 59.88 & 54.82 & 84.21 & 27.45 & 57.65 & 57.95 & 88.79 \\ \hline
        microsoft\_stt  & google\_mt  & \textit{51.19} & \textit{69.31} & \textit{40.74} & 87.87 & 32.98 & 62.92 & 53.77 & 90.07 \\ \hline
        microsoft\_stt  & microsoft\_mt  & 42.67 & 65.03 & 47.43 & 86.64 & \textit{34.47} & 62.75 & 53.37 & 89.37 \\ \hline
        microsoft\_stt  & nllb-3.3B  & 36.48 & 57.72 & 55.78 & 84.02 & 24.81 & 55.71 & 62.66 & 88.62 \\ \hline
        meta\_mms  & ft\_nllb-1.3B  & 32.70 & 55.93 & 59.08 & 79.42 & 24.88 & 56.14 & 59.86 & 86.81 \\ \hline
        meta\_mms  & google\_mt  & 41.92 & 65.07 & 49.33 & 83.20 & 23.41 & 59.35 & 61.41 & 86.94 \\ \hline
        meta\_mms  & microsoft\_mt  & 34.26 & 60.11 & 56.54 & 80.61 & 28.02 & 60.42 & 60.55 & 87.29 \\ \hline
        meta\_mms  & nllb-3.3B  & 31.15 & 53.55 & 61.14 & 79.38 & 20.00 & 53.53 & 65.78 & 86.40 \\ \hline
        whisper  & ft\_nllb-1.3B  & 40.16 & 62.15 & 52.36 & 86.14 & 28.64 & 58.95 & 56.03 & 89.62 \\ \hline
        whisper  & google\_mt  & \textbf{54.27} & \textbf{71.57} & \textbf{37.52} & \textbf{89.60} & 34.27 & \textit{63.96} & \textit{52.46} & \textbf{90.93} \\ \hline
        whisper  & microsoft\_mt  & 44.97 & 67.07 & 44.67 & \textit{88.42} & \textit{36.02} & \textit{64.02} & \textit{51.43} & \textit{90.48} \\ \hline
        whisper  & nllb-3.3B  & 38.46 & 59.54 & 53.69 & 85.92 & 25.76 & 56.65 & 61.41 & 89.37 \\ \hline
    \end{tabular}
    \caption{MT scores of all tested combinations of modules for Turkish and French (both directions).  The two best scores in each column are in \textbf{bold} and the next two in \textit{italics}.  The pipelines are ordered alphabetically by name of ASR and then of MT.}
  \label{tab:fr-and-tr-raw-MT}
\end{table*}

\begin{table*}[ht]
\setlength{\tabcolsep}{4pt}
    \centering
    \begin{tabular}{ll|cccc|cccc}
    \hline
        \textbf{} & \textbf{} & \multicolumn{4}{c|}{\bf ps-fr} & \multicolumn{4}{c}{\bf fr-ps} \\ \hline
        \textbf{ASR} & \textbf{MT} & \textbf{BLEU} & \textbf{ChrF} & \textbf{TER} & \textbf{COMET} & \textbf{BLEU} & \textbf{ChrF} & \textbf{TER} & \textbf{COMET} \\ \hline
        ft\_whisper & google\_mt & - & - & - & - & \textbf{64.22} & \textbf{76.06} & \textbf{29.78} & \textbf{87.03}  \\ \hline
        ft\_whisper & microsoft\_mt & - & - & - & - & 20.49 & 43.92 & 69.42 & 76.69 \\ \hline
        google\_stt & google\_mt & 4.23 & 18.38 & 88.81 & 54.02 & 44.16 & 63.36 & 42.97 & 82.21 \\ \hline
        google\_stt & microsoft\_mt & 6.61 & 20.69 & 87.89 & 54.26 & 19.61 & 41.97 & 72.91 & 74.30 \\ \hline
        meta\_mms & google\_mt & - & - & - & - & 42.17 & 61.40 & 44.77 & 81.56 \\ \hline
        meta\_mms & microsoft\_mt & - & - & - & - & 18.67 & 40.70 & 73.16 & 73.20\\ \hline
        microsoft\_stt & google\_mt & \textbf{25.96} & \textbf{47.43} & \textbf{64.43} & \textbf{77.50} & 56.37 & 70.51 & 36.16 & 84.30 \\ \hline
        microsoft\_stt & microsoft\_mt & \textit{21.36} & \textit{42.87} & \textit{70.84} & \textit{75.13} & 19.67 & 42.83 & 70.06 & 75.66 \\ \hline
        whisper & google\_mt & 9.11 & 27.85 & 90.45 & 57.21 & \textit{56.87} & \textit{71.81} & \textit{35.23} & \textit{85.82} \\ \hline
        whisper & microsoft\_mt & 8.39 & 27.07 & 91.60 & 55.09 & 19.44 & 43.27 & 70.35 & 76.22 \\ \hline
    \end{tabular}
    \caption{MT scores of all tested combinations of modules for Pashto and French (both directions).  The best score in each column is in \textbf{bold} and the second one in \textit{italics}.  The pipelines are ordered alphabetically by name of ASR and then of MT. The ASR system from Meta does not support Pashto, and we did not have enough data to fine-tune Whisper for Pashto.}
  \label{tab:fr-and-ps-raw-MT}
\end{table*}

\end{document}